\documentclass{article}

\usepackage[preprint]{neurips_2026}

\usepackage[utf8]{inputenc} %
\usepackage[T1]{fontenc}    %
\usepackage{hyperref}       %
\usepackage{url}            %
\usepackage{booktabs}       %
\usepackage{amsfonts}       %
\usepackage{nicefrac}       %
\usepackage{microtype}      %
\usepackage{xcolor}         %

\usepackage{amsmath}
\usepackage{amssymb}
\usepackage{mathtools}
\usepackage{tikz-cd}
\usepackage{multirow}
\usepackage[inline]{enumitem}
\usepackage{wrapfig}
\usepackage[most]{tcolorbox}

\newtcolorbox{sectionbox}{
    colframe=black,
    boxrule=0.5pt,
    sharp corners,
    top=6pt, bottom=6pt,
    left=5pt, right=5pt
}

\DeclareMathOperator{\E}{\mathbb{E}}
\DeclareMathOperator{\Var}{\mathbb{V}ar}

\definecolor{customRed}{HTML}{F25D64}
\definecolor{customGreen}{HTML}{63A6A0}
\definecolor{customBlue}{HTML}{08519C}
\def \red {\tikz\draw[customRed,fill=customRed] (0,0) circle (.8ex); }

\def \blue {\tikz\draw[customBlue,fill=customBlue] (0,0) circle (.8ex); }

\def \redsquare    {\tikz\draw[customRed,fill=customRed]     (-0.8ex,-0.8ex) rectangle (0.8ex,0.8ex); }
\def \bluesquare   {\tikz\draw[customBlue,fill=customBlue]   (-0.8ex,-0.8ex) rectangle (0.8ex,0.8ex); }
\def \greensquare  {\tikz\draw[customGreen,fill=customGreen] (-0.8ex,-0.8ex) rectangle (0.8ex,0.8ex); }

\def \redcircle    {\tikz\draw[customRed,fill=customRed]     (0,0) circle (0.8ex); }
\def \bluecircle   {\tikz\draw[customBlue,fill=customBlue]   (0,0) circle (0.8ex); }
\def \greencircle  {\tikz\draw[customGreen,fill=customGreen] (0,0) circle (0.8ex); }

\def \redtriangle  {\tikz\draw[customRed,fill=customRed]     (-0.9ex,-0.8ex) -- (0.9ex,-0.8ex) -- (0,0.8ex) -- cycle; }
\def \bluetriangle {\tikz\draw[customBlue,fill=customBlue]   (-0.9ex,-0.8ex) -- (0.9ex,-0.8ex) -- (0,0.8ex) -- cycle; }

\usepackage{mdframed}
\newmdenv[
  linecolor=black,
  linewidth=0.5pt,
  font=\ttfamily\footnotesize,
]{taskinstructions}

\usepackage{titlesec}
\titlespacing*{\paragraph}{0pt}{0.5ex plus 0.2ex minus 0.1ex}{1em}

\title{Thought without systematicity?\\Evaluating reasoning models on rule induction tasks}

\author{%
  Simon Schug \\
  Princeton University\\
  \texttt{sschug@princeton.edu}\\
  \And
  Brenden M. Lake \\
  Princeton University\\
  \texttt{brenden@princeton.edu}\\
}

\begin{document}

\maketitle

\begin{abstract}
A central tenet of human cognition is systematicity, the principle that understanding one concept is inherently tied to understanding close variations of that concept.
Do reasoning models robustly exhibit such systematicity?
If so, we would expect consistent performance on structurally equivalent variants of the same task.
Here, we extend established rule induction tasks from cognitive science to assess the systematicity of thought in current reasoning models.
Each task family has compositional structure that we use to create structurally equivalent task variations through task isomorphisms such as recombination and substitution.
We find that despite being able to correctly solve a task, models often fail on structurally equivalent variants of the same task.
These findings suggest that many model behaviors lack systematicity, rendering it difficult to robustly establish the cognitive abilities of reasoning models beyond the particular contexts they were evaluated in.
\end{abstract}

\vspace{-1em}
\begin{center}
    \small
    \textbf{Code: } \url{https://github.com/smonsays/systematicity-eval}
\end{center}

\begin{figure}[h]
\begin{center}
  \includegraphics[width=\textwidth]{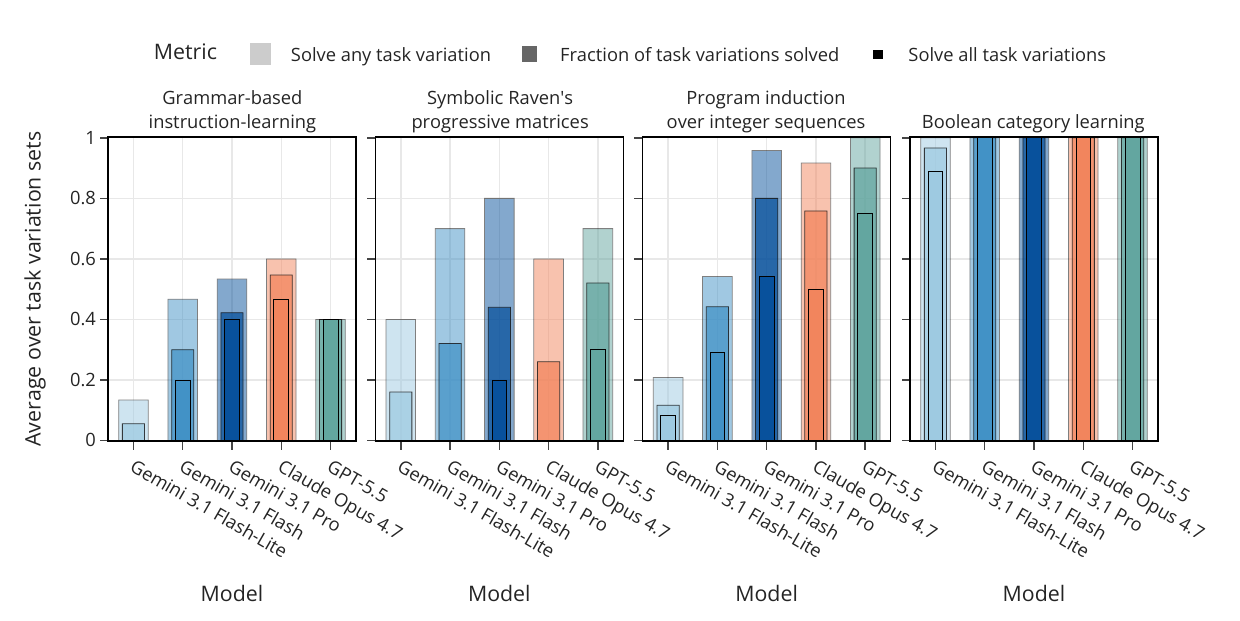}
  \vspace{-2em}
  \caption{\textbf{Systematicity of reasoning models on rule induction tasks.} We evaluate reasoning models on compositional families of rule induction tasks. For each particular task within a family, we generate $K=5$ structurally equivalent task variants per type of task invariance. 
  For each set of task variations, we report whether any of the variations was solved correctly, the fraction of variations solved correctly and whether all variations were solved correctly and report the average over task variation sets within a task family. We find that almost all models are perfectly systematic on the Boolean category learning task family, but see a notable systematicity gap between the ability to solve any task variation and solving all task variations on the remaining task families.}
  \label{fig:systematicity-gap}
  \vspace{-2em}
\end{center}
\end{figure}

\section{Introduction}
\begin{figure}[t]
    \centering
    \includegraphics[width=\linewidth]{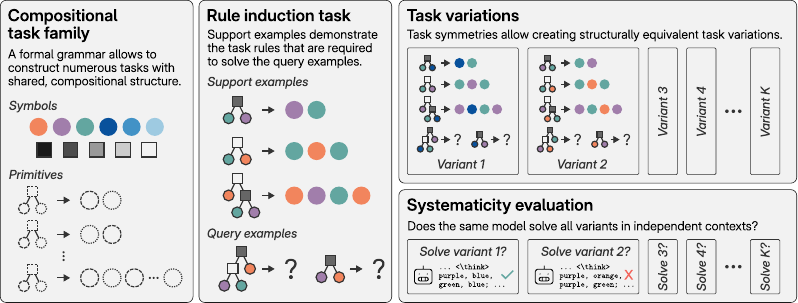}
    \caption{\textbf{Evaluating systematicity in rule induction tasks with compositional structure.} (A) Compositional task families allow to construct large numbers of tasks with related structure. The shown task is a simplified version of the grammar-based instruction-learning task presented in Section~\ref{sec:mlc}. (B)~ Each rule induction task requires inferring hidden rules on a set of input$\rightarrow$output support examples to solve novel query examples. (C)~The compositional task structure allows creating systematic variations of structurally equivalent tasks. (D)~Reasoning models can be independently evaluated across structurally equivalent task variants to assess their systematicity.
    }
    \label{fig:overview}
\end{figure}

Systematicity is a defining feature of human language and thought:
When we can understand one concept it implies that we can also understand many closely related concepts \citep{fodor_connectionism_1988, mclaughlin_connectionismclassicism_1993}.
For instance, when we learn how to \textit{blicket}, we can also \textit{blicket} twice, and if we are able to learn that all red squares are \textit{wudsy} we are equally able to learn that all blue circles are \textit{wudsy}.
This systematicity reflects humans' flexibility to effortlessly generate and comprehend unseen variations of familiar elements \citep{chomsky_syntactic_1985,lake_building_2017} and it is a crucial assumption underlying cognitive tests, because it enables measuring an ability in a particular context but make inferences about that ability beyond the specific context within which it was measured.

The systematicity of human cognition has many facets reflecting the various ways in which a task can be altered according to its underlying compositional structure \citep{hupkes_compositionality_2020}.
This includes substitutions such as using a different symbol for the same underlying concept, or recombinations of the parts of an expression that serve similar roles.
These invariances reflect compositional task structure relating tasks to each other through a common syntax that prescribes how sets of primitives can be composed into different configurations.
From this perspective, systematicity can be understood as a behavioral outcome of successfully capturing the underlying compositional structure of a family of tasks.
Compositional task families are exponential in nature and it is virtually impossible to exhaustively include all possible task variations in the training data even at large scale \citep{schug_meta-learning_2025}.
This raises the question: \textit{Do reasoning models exhibit systematicity of thought and perform consistently across the many possible, structurally equivalent variations of a given task?}

Here, we study this question by drawing on established experimental paradigms from cognitive science.
In particular, we consider rule induction tasks that require the learner to infer latent structure, and evaluate it across structurally equivalent task variations to probe the systematicity of its thinking as illustrated in Figure~\ref{fig:overview}.
A variety of such tasks have been studied in cognitive science, ranging from simple Boolean concept learning, abstract reasoning over symbolic sequences, to discovering the syntax and semantics of artificial languages \citep{raven_advanced_1962, feldman_minimization_2000,piantadosi_logical_2016,lake_human-like_2023,rule_symbolic_2024}.
These tasks have compositional structure that allows us to generate a large number of possible tasks and probe understanding beyond the specifics of any particular experience.
Importantly, for each specific task we can create structurally equivalent task variations by using \textit{task isomorphisms} -- equivariant task transformations based on the underlying compositional structure -- to evaluate systematicity.
Unlike humans whose mental states are invariably altered by experience, models provide the opportunity to evaluate systematicity in its strictest form by observing behavioral outcomes in multiple counterfactual worlds that do not affect each other.

\begin{sectionbox}
We summarize our \textbf{main contributions} in the following:
\begin{itemize}[leftmargin=*]
  \item In Section~\ref{sec:methods}, we formalize systematicity and how to measure it over structurally equivalent rule induction tasks via task isomorphisms.
  \item In Section~\ref{sec:task-families}, we adapt four rule induction task families for generating systematic task variations.
  \begin{itemize}
      \item Section~\ref{sec:mlc}: Grammar-based instruction learning \citep{lake_human-like_2023}
      \item Section~\ref{sec:raven}: A symbolic version of Raven's progressive matrices \citep{schug_attention_2025}
      \item Section~\ref{sec:listint}: Program induction over integer sequences \citep{rule_symbolic_2024}
      \item Section~\ref{sec:boolean}: Boolean category learning \citep{piantadosi_logical_2016}
  \end{itemize}
  \item In Section~\ref{sec:systematicity-evaluation}, we evaluate the systematicity of reasoning models across task variations on our four task families and investigate the impact of reasoning effort and sampling noise.
\end{itemize}
\end{sectionbox}

\section{Related work}
The observation that human cognition is systematic has been a longstanding locus of inquiry, reaching back to early attempts to formulate laws of thought \citep{boole_investigation_1854,griffiths_laws_2026}.
Systematicity is a central tenet of the language of thought hypothesis -- a prominent theory of human cognition which assumes that mental representations have combinatorial syntax and semantics \citep{fodor_language_1975, fodor_connectionism_1988,penn_darwins_2008, goodman_concepts_2015}.
Cognitive science therefore has a rich history of developing tests of rule learning, with varying latent syntactic and semantic structure, to investigate the language of thought hypothesis.
Classic work by \citet{feldman_minimization_2000} uses Boolean concept learning tasks to show that subjective difficulty is directly related to the complexity of the underlying rule and learning tasks with more complex logical rules have been used to investigate what concrete primitives might underlie the language of thought \citep{piantadosi_logical_2016, rule_symbolic_2024}.
Tasks that require inferring compositional string rewriting rules have further been used to compare compositional learning between humans and neural networks, a challenge neural networks have historically struggled with but recently made notable progress on \citep{lake_human-like_2023}.

Despite this progress, early large language models were brittle and could be derailed by subtle variations in their inputs, including typos, irrelevant sentences, changes in phrasing,  and reordering of examples \citep{jia_adversarial_2017, jiang_how_2020, zhao_calibrate_2021, lu_fantastically_2022}.
While instruction tuning via human feedback \citep{ouyang_training_2022}, progressive scaling \citep{kaplan_scaling_2020, bubeck_sparks_2023} and inference-time reasoning \citep{wei_chain--thought_2022, wang_self-consistency_2022, lightman_let_2024,openai_openai_2024} have rendered the resulting models significantly less sensitive to surface changes, adversarial brittleness remains a central concern \citep{wang_decodingtrust_2023, zou_universal_2023, berglund_reversal_2024, romanou_brittlebench_2026, mondorf_lpds_2026, burnell_measuring_2026}, in particular in the context of safety and alignment \citep{wei_jailbroken_2023, chen_sage-eval_2025}.

In recent years, stronger emphasis has therefore been put on evaluating model consistency through metamorphic testing where a known relationship between outputs of related inputs is evaluated \citep{segura_survey_2016}.
This includes consistency across languages \citep{cho_metamorphic_2025} or category-based substitution \citep{ribeiro_beyond_2020}.
Logical consistency testing in particular evaluates the internal consistency of model knowledge \citep{jang_becel_2022}, e.g. by verifying invariance to paraphrasing \citep{elazar_measuring_2021} or whether predicted relations are closed under transitivity \citep{jang_consistency_2023}.
Our systematicity evaluation can be interpreted as a specific type of metamorphic test that verifies invariant task-solving ability with respect to a compositional data generating procedure.

\section{What is systematicity and how can we measure it?}
\label{sec:methods}
In the following, we formalize systematicity, how it can be measured in the setting of rule induction tasks and develop metrics to quantify a learner's systematicity on such tasks.

\subsection{What is systematicity?}
We define systematicity as invariance of ability with respect to certain transformations of a task with compositional structure.
For example, we might expect the ability to understand an expression like "Alice loves John" to be invariant to a permutation of its constituents like "John loves Alice".
This means we treat systematicity as a latent property of behavior that makes assumptions about the compositional structure underlying a data generating process.
To quantify systematicity, we will measure how the ability to solve a task varies across task variations that modify the constituents of a task according to its compositional structure.
Since some task variations might change the difficulty of a task and confound the systematicity measure, we will restrict task variations to be structurally equivalent to each other through isomorphisms, one-to-one mappings between each task variation.

\subsection{Types of systematic task variations}
The particular ways in which task constituents can be modified depend on the compositional structure of the data generating process.
In our task families we will consider the following task transformations:
\begin{itemize}[leftmargin=*]
  \item \textbf{Symbol substitution}: Replacing symbols that carry no intrinsic semantic meaning relevant to the task. For example replacing one pseudoword "dax" with another pseudoword "lug".
  \item \textbf{Feature rebinding}: Rebinding latent variables to different features. For example, representing the same number through size, orientation or numerosity.
  \item \textbf{Constituent permutation}: Shuffling constituents of the same type within an expression. For example, "Alice loves John" $\leftrightarrow$ "John loves Alice".
  \item \textbf{Example reordering}: Changing the order with which multiple independent examples are presented.
\end{itemize}

\subsection{Structurally equivalent rule induction tasks}
We will evaluate systematicity within the setting of rule induction tasks as illustrated in Figure~\ref{fig:overview}.
In each task, we present a learner with a set of input-output examples based on which it has to infer hidden underlying rules.
We then verify whether the learner correctly inferred the rules by asking it to predict the outputs on a novel set of inputs.
For instance, in the Boolean category learning task, a learner might be presented with three support examples,
$\redsquare \rightarrow \texttt{wudsy}$, $\bluesquare \rightarrow \texttt{not wudsy}$, $ 
\redtriangle \rightarrow \texttt{wudsy}$,
from which it has to infer the simplest rule that explains which objects are \texttt{wudsy} (\textit{all red objects are wudsy}).
We then evaluate whether it did so correctly by asking it to complete query examples such as, $\greensquare \rightarrow \textbf{?}$ and $\redcircle \rightarrow \textbf{?}$.

To study a learner's systematicity across variations of a task of equal difficulty, we construct structurally equivalent tasks through \textit{isomorphisms}.
An isomorphism is a structure-preserving mapping that is invertible and can therefore only alter the surface characteristics of a task but not its underlying structure.
For example, we can construct the structurally equivalent task $\bluecircle \rightarrow \texttt{wudsy}$, $\greencircle \rightarrow \texttt{not wudsy}$, $ 
\bluesquare \rightarrow \texttt{wudsy}$, $\redcircle \rightarrow \textbf{?}$ and $\bluetriangle \rightarrow \textbf{?}$, where the hidden rule is \textit{all blue objects are wudsy} and the colors and shapes were remapped accordingly.

\subsection{Isomorphic task variations}
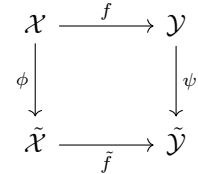
\begin{wrapfigure}{R}{0.20\textwidth}
    \vspace{-2em}
    \centering
    \begin{tikzcd}[row sep=large, column sep=large]
        \mathcal{X} \arrow[r, "f"] \arrow[d, "\phi"'] & \mathcal{Y} \arrow[d, "\psi"] \\
        \tilde{\mathcal{X}} \arrow[r, "\tilde{f}"'] & \tilde{\mathcal{Y}}
    \end{tikzcd}
    \caption{Relation-ship between the original and transformed hidden rule.}
    \label{fig:commutative-diagram-isomorphism}
    \vspace{-2em}
\end{wrapfigure}

Formally, let $\mathcal{X}$ and $\mathcal{Y}$ be input and output spaces. 
We define a task as a tuple $T = (f, S, Q)$, where $f: \mathcal{X} \to \mathcal{Y}$ is a \textit{hidden rule}, $S = \left( (x_i, f(x_i)) \right)_{i=1}^N$ is the \textit{support set}, and $Q = (x'_j)_{j=1}^M$ is the \textit{query set}, with $x_i, x'_j \in \mathcal{X}$.
Upon observing only the support set $S$, a learner is tasked to predict the targets of the query set, $f(x'_j)$ for all $j=1, \dots, M$.
We say a learner solves a task if it correctly predicts the whole query set.

An \textit{isomorphic task variation} is generated by applying a transformation $\tau = (\pi, \sigma, \phi, \psi)$, where $\pi$ and $\sigma$ are permutations of the indices $\{1, \dots, N\}$ and $\{1, \dots, M\}$ respectively, and $\phi: \mathcal{X} \to \tilde{\mathcal{X}}$ and $\psi: \mathcal{Y} \to \tilde{\mathcal{Y}}$ are bijections that map elements from the original input and output space to new input and output spaces $\tilde{\mathcal{X}}$ and $\tilde{\mathcal{Y}}$.
Applying $\tau$ yields a transformed task $\tau(T) = (\tilde{f}, \tilde{S}, \tilde{Q})$, where the new hidden rule evaluates as $\tilde{f} = \psi \circ f \circ \phi^{-1}$, the transformed and permuted support set is $\tilde{S} = \left( \phi(x_{\pi(i)}), \psi(f(x_{\pi(i)})) \right)_{i=1}^N$, and the transformed and permuted query set is $\tilde{Q} = \left( \phi(x'_{\sigma(j)}) \right)_{j=1}^M$.
\vspace{1em}
\subsection{Systematicity metrics}
\label{sec:systematicity-metrics}

\begin{wrapfigure}{R}{0.27\textwidth}
\vspace{-1em}
  \centering
  \includegraphics[width=0.9\linewidth]{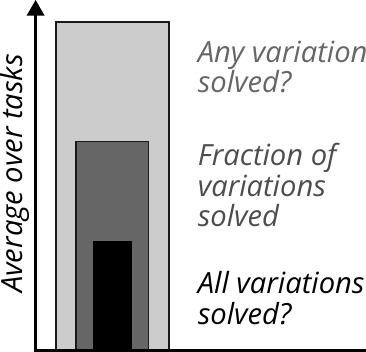}
  \caption{We use bullet charts to compactly visualize the three systematicity metrics.}
  \label{fig:systematicity-bullet}
  \vspace{-6em}
\end{wrapfigure}
A strictly systematic learner should be invariant to isomorphic task variations assuming that the semantics of any two input spaces are equally (un)informative for solving the task:
If it can grasp the underlying structure well enough to solve one task variation, we would expect it to be able to solve them all.
Let $\operatorname{pass}(T) \in \{0, 1\}$ be a binary indicator denoting whether a learner correctly predicts the targets for the entire query set of task $T$.
To evaluate a learner's robustness to isomorphic variations, we sample $K$ task transformations $\{\tau_1, \dots, \tau_K\}, \tau_k \sim p(\tau)$ for each task $T$  and calculate the following systematicity metrics:
\begin{itemize}[leftmargin=*]
  \item \textbf{Solve any of $K$ variations}: $\max_{k \in \{1, \dots, K\}} \operatorname{pass}(\tau_k(T))$
  \item \textbf{Fraction of variations solved}: $\frac{1}{K} \sum_{k=1}^K \operatorname{pass}(\tau_k(T))$
  \item \textbf{Solve all $K$ variations}: $\min_{k \in \{1, \dots, K\}} \operatorname{pass}(\tau_k(T))$
\end{itemize}

Collectively, these three metrics allow us to characterize the systematicity with which a learner solves a task:
If it never passes any task variation, this means the tasks are too difficult for the learner.
If it passes some task variations but not all of them, it is not fully systematic and the fraction of variations passed captures to what extent.
If all task variations are passed it can be considered systematic on this task with respect to the tested task variations.
We can take advantage of the fact that the three metrics are monotonically decreasing and visualize them as a bullet chart as shown in Figure~\ref{fig:systematicity-bullet}.

\subsection{Systematicity under stochasticity}
\label{sec:model-stochasticity}

An important consideration in the context of systematicity is how to handle possible stochasticity in the learner that solves the tasks.
When repeatedly evaluating a stochastic learner on a given task, the learner might randomly fail to complete the task on some attempts.
While large language models were traditionally evaluated using greedy decoding, current reasoning models have been found to perform better when evaluated with stochastic sampling \citep{wang_self-consistency_2022, guo_deepseek-r1_2025}.
In fact, many proprietary model providers now only allow to use reasoning models with a positive sampling temperature.
To account for the resulting decoding variance, we give each reasoning model multiple attempts per task variation (here we use $N=5$ attempts throughout) and consider a variation as passed if it was correctly solved in the majority of the attempts.
We will further study the impact of stochasticity on systematicity in Section~\ref{sec:results-stochasticity}. 

\section{Rule induction tasks for evaluating systematicity}
\label{sec:task-families}

In the following sections, we present four families of rule induction tasks built on established task paradigms from cognitive science.
Each data generating procedure relies on some form of compositional structure that allows to sample a large number of compositional tasks of varying difficulty.
Their synthetic nature provides us with the necessary control to create task variations that are guaranteed to be isomorphic in order to evaluate systematicity.

\subsection{Grammar-based instruction-learning}
\label{sec:mlc}
The grammar-based instruction-learning task family was introduced by \citet{lake_human-like_2023}.
In each task from this family, the model must infer the latent grammatical rules of an artificial language from a few demonstrations to translate pseudolanguage commands into a sequence of outputs.
Since the artificial languages are procedurally generated from a meta-grammar, an infinite number of such languages of varying complexity can in principle be generated.
This task family was originally designed to evaluate the ability of humans and neural networks for compositional generalization, the ability to solve unseen task compositions made from familiar parts.

\paragraph{Instructions}
To reduce the influence of prior experience on task performance as well as limit possible ambiguity, we provide detailed instructions on the general structure of each task in the system prompt, shown in Figure~\ref{fig:mlc-instructions}.

\paragraph{Example}
The following is a sample episode from the grammar-based instruction-learning task family.
\begin{align*}
\texttt{dax fep} &\rightarrow \red\ \red\ \red \\ %
\texttt{lug fep} &\rightarrow \blue\ \blue\ \blue \\ %
\texttt{dax zup lug} &\rightarrow \blue\ \red \\ %
\texttt{dax zup lug fep} &\rightarrow \blue\ \blue\ \blue\ \red \\ %
\texttt{dax fep zup lug} &\rightarrow \blue\ \red\ \red\ \red\ \\ %
\texttt{lug fep zup dax fep} &\rightarrow \textbf{?}
\end{align*}

To predict the single query example of this task, the learner must infer primitive rules, mapping input tokens, like \texttt{dax, lug}, to output tokens, \texttt{RED, BLUE}, and function rules that transform their inputs, like \texttt{fep, zup}. Here \texttt{fep} triples its input and \texttt{zup} swaps its inputs.
The learner then has to apply the inferred rules to a novel input composition and predict its answer, here $\red \, \red \, \red \, \blue \, \blue \, \blue$.

\paragraph{Task structure} 
Each generated language is governed by a uniquely sampled syntax that consists of primitive rules and function rules.
Whereas primitive rules are simple one-to-one mappings between input tokens (e.g., pseudowords like \texttt{dax}, \texttt{lug}) and output tokens (e.g., capitalized colors like \texttt{RED}, \texttt{BLUE}), function rules specify how function tokens transform one or two adjacent arguments into a sequence of outputs.
The function arguments are either a single primitive token or match a whole preceding or succeeding string.
When applied, function rules deterministically reorder, delete or duplicate their arguments to produce a new output sequence. 
To resolve syntactic ambiguities, functions that accept string arguments have a strict precedence order.
In addition, the parser evaluates sequences via left-to-right reduction, ensuring that for any valid input string there exists a unique translation.

\paragraph{Support and query set}
In the original grammar-based instruction-learning task family as used in \citet{lake_human-like_2023}, tasks were hand selected to ensure that their underlying grammar could be unambiguously inferred.
Since we would like to generate a large number of unambiguous tasks, we define a procedure to automatically create an instructive support set that ensures all primitive rules and function rules can be identified for a given, randomly sampled grammar .
Specifically, we construct the support set by producing lexical anchors that demonstrate the primitive rules (e.g., \texttt{dax fep} $\rightarrow$ \red\ \red\ \red\ and \texttt{lug fep} $\rightarrow$ \blue\ \blue\ \blue in the example above),
template resolvers that reveal the arity and output transformations of the rule functions (e.g., \texttt{dax zup lug} $\rightarrow$ \blue\ \red  shows that \texttt{zup} is a two argument reversal function), and precedence proofs that disambiguate the precedence order of multiple string matching function rules.
The query set as well as a configurable number of additional support examples is then generated from nested compositions that involve at least two function rule applications per example, ensuring that there are no duplicate examples across the support and query set.

\paragraph{Task invariances}

The resulting tasks have several invariances that we can use to generate isomorphic task variations.
We can permute the order of both the support and query examples (\textit{example reordering}), apply a bijective mapping to the input and output vocabularies (\textit{symbol substitution}) and permute the primitive tokens in a given expression, e.g. by changing \texttt{dax zup lug} to \texttt{lug zup dax} (\textit{recomposition}).

\subsection{Symbolic Raven's progressive matrices}
\label{sec:raven}
Raven's progressive matrices is a classic human intelligence test \citep{raven_advanced_1962}.
In its original form, each task consists of a three by three grid of abstract symbols with a missing final panel whose contents need to be inferred.
\citet{schug_attention_2025} introduce a symbolic variant of this task family that allows to procedurally generate Raven-like tasks of varying difficulty, creating challenging abstract reasoning problems.

\paragraph{Instructions}
Similar to before we provide detailed instructions on the general structure of tasks from this task family in the system prompt, shown in Figure~\ref{fig:raven-instructions}.

\paragraph{Example}
The following is a sample episode from the symbolic Raven's progressive matrices task family using two  ($F=2$) and modulo 10 arithmetic.
The left side shows the unpermuted base task, while the right side additionally considers a column-specific feature permutation which models the difficulty of \textit{finding correspondences} \citep{carpenter_what_1990}.
\begin{align*}
&\begin{array}{ccc}
\begin{bmatrix} 5 \\ 2 \end{bmatrix} & \begin{bmatrix} 5 \\ 4 \end{bmatrix} & \begin{bmatrix} 5 \\ 6 \end{bmatrix} \\[2em]
\begin{bmatrix} 8 \\ 7 \end{bmatrix} & \begin{bmatrix} 8 \\ 9 \end{bmatrix} & \begin{bmatrix} 8 \\ 1 \end{bmatrix} \\[2em]
\begin{bmatrix} 3 \\ 5 \end{bmatrix} & \begin{bmatrix} 3 \\ 7 \end{bmatrix} & \begin{bmatrix} \textbf{?} \\ \textbf{?} \end{bmatrix}
\end{array}
\quad \Leftrightarrow \quad
\begin{array}{ccc}
\begin{bmatrix} 2 \\ 5 \end{bmatrix} & \begin{bmatrix} 5 \\ 4 \end{bmatrix} & \begin{bmatrix} 6 \\ 5 \end{bmatrix} \\[2em]
\begin{bmatrix} 7 \\ 8 \end{bmatrix} & \begin{bmatrix} 8 \\ 9 \end{bmatrix} & \begin{bmatrix} 1 \\ 8 \end{bmatrix} \\[2em]
\begin{bmatrix} 5 \\ 3 \end{bmatrix} & \begin{bmatrix} 3 \\ 7 \end{bmatrix} & \begin{bmatrix} \textbf{?} \\ \textbf{?} \end{bmatrix}
\end{array}
\end{align*}
To solve the unpermuted task (left), the learner must infer the rules applied horizontally to each aligned feature dimension.
Here, the first feature is constant (\texttt{8} $\rightarrow$ \texttt{8} $\rightarrow$ \texttt{8}), and the second feature follows an arithmetic progression of +2 modulo 10 (e.g., \texttt{7} $\rightarrow$ \texttt{9} $\rightarrow$ \texttt{1}).
Applying these rules to the third row we obtain the target panel, $\begin{bmatrix} 3 \\ 9 \end{bmatrix}$.

In the permuted variant (right), the task is significantly more difficult because the features are no longer spatially aligned across columns.
The learner must simultaneously discover the latent rules and the implicit feature correspondence -- recognizing, for example, that the constant feature corresponds to the bottom element of column 1, the top element of column 2, and the bottom element of column 3.
After disentangling these mappings and applying the latent rules, the learner must predict the appropriately permuted final panel, $\begin{bmatrix} 9 \\ 3 \end{bmatrix}$.

\paragraph{Task structure}
Each task is structured as a $3 \times 3$ grid of panels, where each panel contains an $F$-dimensional feature vector of integers, governed by a hidden combination of $F$ rules and ordered according to a column-specific permutation.
The rules operate horizontally, such that the features of the third column in each row are determined by applying independent rules to the features in the first two columns.
The same sequence of rules is applied consistently across all three rows.
The available rules encompass constant patterns, arithmetic progressions, modular addition or subtraction, minimum/maximum operations, and the distribution of distinct elements.

\paragraph{Support and query set}
To generate a task, we sample a hidden combination of rules, one for each feature dimension as well as column-specific feature permutations.
The first two rows of the $3 \times 3$ matrix act as the support set, demonstrating the applied rules, while the first two columns of the third row serve as the query.

\paragraph{Task invariances}
The column-specific feature permutations can be used to create isomorphic transformations of the same task, rebinding the hidden features of the unpermuted task to the observed feature orderings (\textit{feature rebinding}).

\subsection{Program induction over integer sequences}
\label{sec:listint}
Next, we consider rule induction over integer sequences as studied by \citet{rule_symbolic_2024} in which a learner must infer a latent rule to transform an input list of integers into an output list of integers.
Since the original set of tasks were handcrafted, we define a probabilistic context-free grammar that allows us to procedurally sample a large number of similar tasks and systematically create isomorphic task variations.

\paragraph{Instructions}
We provide the general task description shown at the top of Figure~\ref{fig:listint-instructions} to the learner in the main evaluation and perform additional experiments where we additionally describe the basic operations from which each task is composed in the instructions as shown at the bottom of Figure~\ref{fig:listint-instructions}.

\paragraph{Example}
The following is a sample episode from the program induction over integer sequences task family.
\begin{align*}
\texttt{[1, 2, 3, 4]} &\rightarrow \texttt{[4, 1, 2]} \\
\texttt{[1, 8, 3]} &\rightarrow \texttt{[3, 1]} \\
\texttt{[8, 9, 1, 2, 4]} &\rightarrow \texttt{[2, 4, 8, 9]} \\
\texttt{[0, 1]} &\rightarrow \texttt{[0]} \\
\texttt{[3, 3, 4, 5, 5, 6]} &\rightarrow \texttt{[5, 6, 3, 3, 4]} \\
\texttt{[2]} &\rightarrow \texttt{[]} \\
\texttt{[7, 3, 1, 5, 2]} &\rightarrow \textbf{?}
\end{align*}
To predict the query in this example, the learner must infer that the target rule is a composition of two operations: first swapping the two halves of the list, and then removing the element at the first position.
It then has to apply these operations to the novel test input sequence to predict the answer, here \texttt{[5, 2, 7, 3]}.

\paragraph{Task structure} 
The underlying program for each task is sampled from a probabilistic context-free grammar that allows to create compositions of operations that take integer lists as input and output integer lists.
The grammar supports parameterized operations such as \texttt{insert} or \texttt{remove}, which take additional integer arguments such as structural positions, sequence lengths, or vocabulary elements.
The full list of operations is shown at the bottom of Figure~\ref{fig:listint-instructions}.
By composing these operations, the generative model produces a single deterministic expression that parses and transforms any given input list.
Parameters and input sequence integers are sampled uniformly from predefined ranges.

\paragraph{Support and query set}
We randomly sample input lists of integers in the range of zero to nine for the support and query set, strictly partitioning the sequence lengths of the support and query sets into disjoint pools.
Specifically, we ensure that the support set contains lists that are both shorter and longer than the lists in the query set.
The query set examples are subsequently sampled from intermediate sequence lengths that are never seen in the support set.
This is to prevent the existence of a shortcut solution in which -- rather than inferring the underlying program -- it would be possible to simply copy the analogical elements of the same length example. 

\paragraph{Task invariances}
We can create isomorphic task variations by permuting the order of the support and query samples (\textit{example reordering}).
In addition, we can apply a bijective mapping to the integer vocabulary of the elements in the support and query set as well as implicitly to any integer parameters in the hidden rule, e.g., the value argument inserted by the \texttt{insert} operation (\textit{symbol substitution}). 

\subsection{Boolean concept learning}
\label{sec:boolean}

Finally, we adapt the Boolean concept learning task family studied in \citet{feldman_minimization_2000} and \citet{piantadosi_logical_2016}.
In each task, a learner must infer a latent concept from examples to classify objects with varying physical features, e.g. shapes with different colors and sizes into binary categories.

\paragraph{Instructions}

Figure~\ref{fig:boolean-instructions} shows the general task instructions we provide in the system prompt asking models to identify the \textit{simplest} rule to classify the support examples.
What constitutes the \textit{simplest} rule is underspecified without knowledge of the description language.
For this reason, for each task we describe the context-free grammar from which the rule can be generated and with respect to which it is the shortest rule that explains the support set, see Figure~\ref{fig:boolean-instructions} for an example.

\paragraph{Example}
The following is a sample episode from the Boolean concept learning task family.
\begin{align*}
\redsquare &\rightarrow \texttt{wudsy} \\
\bluesquare &\rightarrow \texttt{not wudsy} \\
\redtriangle &\rightarrow \texttt{wudsy} \\
\greensquare &\rightarrow \textbf{?}
\end{align*}
To predict the query example of this task, the learner must infer that the simplest Boolean rule that explains the support set is that \textit{all red objects are wudsy}.

\def \hgredsquare {\tikz\draw[black,thick,fill=customRed] (-0.8ex,-0.8ex) rectangle (0.8ex,0.8ex); }
\def \hgbluesquare {\tikz\draw[black,thick,fill=customBlue] (-0.8ex,-0.8ex) rectangle (0.8ex,0.8ex); }
\def \hggreensquare {\tikz\draw[customGreen,fill=customGreen] (-0.8ex,-0.8ex) rectangle (0.8ex,0.8ex); }

\def \hgredtriangle {\tikz\draw[black,thick,fill=customRed] (-0.9ex,-0.8ex) -- (0.9ex,-0.8ex) -- (0,0.8ex) -- cycle; }
\def \hgbluetriangle {\tikz\draw[black,thick,fill=customBlue] (-0.9ex,-0.8ex) -- (0.9ex,-0.8ex) -- (0,0.8ex) -- cycle; }
\def \hggreentriangle {\tikz\draw[customGreen,fill=customGreen] (-0.9ex,-0.8ex) -- (0.9ex,-0.8ex) -- (0,0.8ex) -- cycle; }

\def \hgredcircle {\tikz\draw[black,thick,fill=customRed] (0,0) circle (.8ex); }
\def \hgbluecircle {\tikz\draw[black,thick,fill=customBlue] (0,0) circle (.8ex); }
\def \hggreencircle {\tikz\draw[customGreen,fill=customGreen] (0,0) circle (.8ex); }

\begin{wrapfigure}{R}{0.25\textwidth}
\centering
\begin{tikzpicture}[scale=1.1, every node/.style={inner sep=2pt}]
\fill[customRed, opacity=0.1, rounded corners] (-0.45, -0.4) rectangle (0.3, 2.4);
\fill[customBlue!50!customGreen, opacity=0.1, rounded corners] (0.6, -0.4) rectangle (2.3, 2.4);

\foreach \y in {0,1,2} {
    \draw[gray, very thin] (1,\y) -- (2,\y);
    \draw[gray, very thin] (0,\y) to[bend left=25] (2,\y);
}
\foreach \x in {0,1,2} {
    \draw[gray, very thin] (\x,0) -- (\x,1);
    \draw[gray, very thin] (\x,1) -- (\x,2);
    \draw[gray, very thin] (\x,0) to[bend right=25] (\x,2);
}

\foreach \y in {0,1,2} {
    \draw[black, line width=1.2pt] (0, \y) -- (1, \y);
}

\draw[dashed, black, line width=1.2pt] (0.45, -0.45) -- (0.4, 2.4);

\node (rs) at (0, 2) {\scalebox{1.5}{\hgredsquare}};
\node (bs) at (1, 2) {\scalebox{1.5}{\hgbluesquare}};
\node (gs) at (2, 2) {\scalebox{1.5}{\hggreensquare}};

\node (rt) at (0, 1) {\scalebox{1.5}{\hgredtriangle}};
\node (bt) at (1, 1) {\scalebox{1.5}{\hgbluetriangle}};
\node (gt) at (2, 1) {\scalebox{1.5}{\hggreentriangle}};

\node (rc) at (0, 0) {\scalebox{1.5}{\hgredcircle}};
\node (bc) at (1, 0) {\scalebox{1.5}{\hgbluecircle}};
\node (gc) at (2, 0) {\scalebox{1.5}{\hggreencircle}};

\node[below] at (-0.05, -0.4) {\footnotesize \texttt{wudsy}};
\node[below] at (1.45, -0.4) {\footnotesize \texttt{not wudsy}};

\end{tikzpicture}
\caption{Example Hamming graph used to construct the support set from objects at the decision boundary (dashed line) of a given rule.}
\label{fig:boolean-hamming-graph}
\vspace{-2em}
\end{wrapfigure}
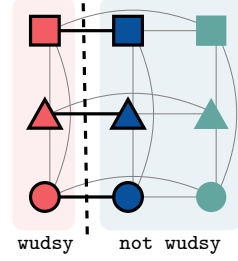

\paragraph{Task structure}
We procedurally generate Boolean logic expressions of varying complexity from a probabilistic context-free grammar.
Our grammar supports generating compositions of the logical operators \texttt{and} and \texttt{or}, as well as base predicates that evaluate specific features of the objects (e.g., \texttt{is\_color(x, red)}.
Notably, the grammar does not contain a \texttt{not} operator to reduce redundancy in the hypothesis space over possible rules.
The particular features objects have in a specific task are a random subset of three out of five possible features (shape, size, color, material, and numerosity), where each feature can take three distinct values (e.g., red, blue, green).

\paragraph{Support and query set}

When naively sampling a support set for a given rule, it is likely that a simpler rule than the one used to label the support examples can equally explain the support set.
To contend with this possibility, we strategically build the support set and use an ideal observer model to ensure the rule used to generate the set is unique and the simplest to explain it.
Specifically, to create the support set, we construct the Hamming graph over all theoretically possible objects, with an edge between any two objects that differ by exactly one feature as shown in Figure~\ref{fig:boolean-hamming-graph}.
As a heuristic, we can then select all pairs of adjacent objects in the Hamming graph that lie on the decision boundary of the given rule and randomly sample the query set from the remaining objects.

\paragraph{Task invariances}

The specific physical feature to which a rule applies should not matter to a systematic learner, i.e. the rule \textit{all red objects are wudsy} is structurally equivalent to \textit{all squares are wudsy}.
We can therefore create isomorphic task variations both by permuting the order of the support and query samples (\textit{example reordering}) as well as by rebinding the physical features of a rule and correspondingly changing the support and query set (\textit{symbol substitution}).

\section{Evaluating reasoning models across systematic task variations}
\label{sec:systematicity-evaluation}

We now evaluate systematicity of current state-of-the-art proprietary reasoning models on the rule induction task families introduced above.
For each task, we generate a set of isomorphic task variations, compute the systematicity metrics defined in Section~\ref{sec:systematicity-metrics} and report the average across task variation sets within a task family.
While ideally we would cover all possible variations, there is typically a prohibitively large number of possible variations.
Throughout our experiments, we therefore sample $K=5$ variations per task isomorphism.

Apart from experiments where we explicitly vary the reasoning effort, we evaluate all models with reasoning effort set to `high' and use the default sampling parameters of the official APIs for each model.
We use structured outputs to ensure models output consistently formatted responses according to the schema requirements of each task family.
We compare models from three major model providers in the main evaluation and focus on Gemini models to examine the effect of reasoning effort to keep costs reasonable.

\subsection{Models can solve some task variations but struggle to systematically solve all variations}

Figure~\ref{fig:systematicity-gap} shows average systematicity metrics over task variation sets on the four rule induction task families.
Across most of the task families, we find that there is a notable systematicity gap, i.e. a clear difference between the fraction of task variation sets where any variation was solved and the fraction of task variation sets where all variations were solved.
The \textit{Boolean category learning} task is a notable exception to this observation.
Almost all models can solve all tasks and variations thereof, demonstrating that perfect systematicity is achievable.

The systematicity gap appears most clearly for \textit{symbolic Raven's progressive matrices} and \textit{program induction over integer sequences}.
For example, in $91.67\%$ of the task variation sets in the program induction task family, Claude Opus 4.7 can solve at least one variation but only in $50.00\%$ of the task variation sets can it solve all variations, leaving a systematicity gap of $41.67\%$.
GPT-5.5 

We can decrease the difficulty of the program induction tasks by providing a detailed \textit{hint} of the possible program components (see Figure~\ref{fig:listint-instructions}) and accordingly observe a notable improvement in performance for stronger models in Figure~\ref{fig:listint} albeit without fully resolving the systematicity gap.
The Raven's matrices task family already uses the easiest parameterization with only two features.
Making it more difficult by increasing the number of features to three predictably further decreases model performance further, as shown in Figure~\ref{fig:raven}.

On the \textit{grammar-based instruction-learning} task family overall performance is modest and models lack systematicity with the notable exception of GPT-5.5: if it can solve any task variation it can solve all of them, otherwise it solves none.
Figure~\ref{fig:mlc}A further demonstrates that the systematicity gap is not specific to a particular type of task isomorphism but appears to varying degrees for recomposition, reordering and substitution.
In addition, Figure~\ref{fig:mlc}B shows that while adding extra support examples alongside the curriculum examples makes the task easier, the systematicity gap remains.

\begin{figure}[t]
\begin{center}
    \includegraphics[width=\textwidth]{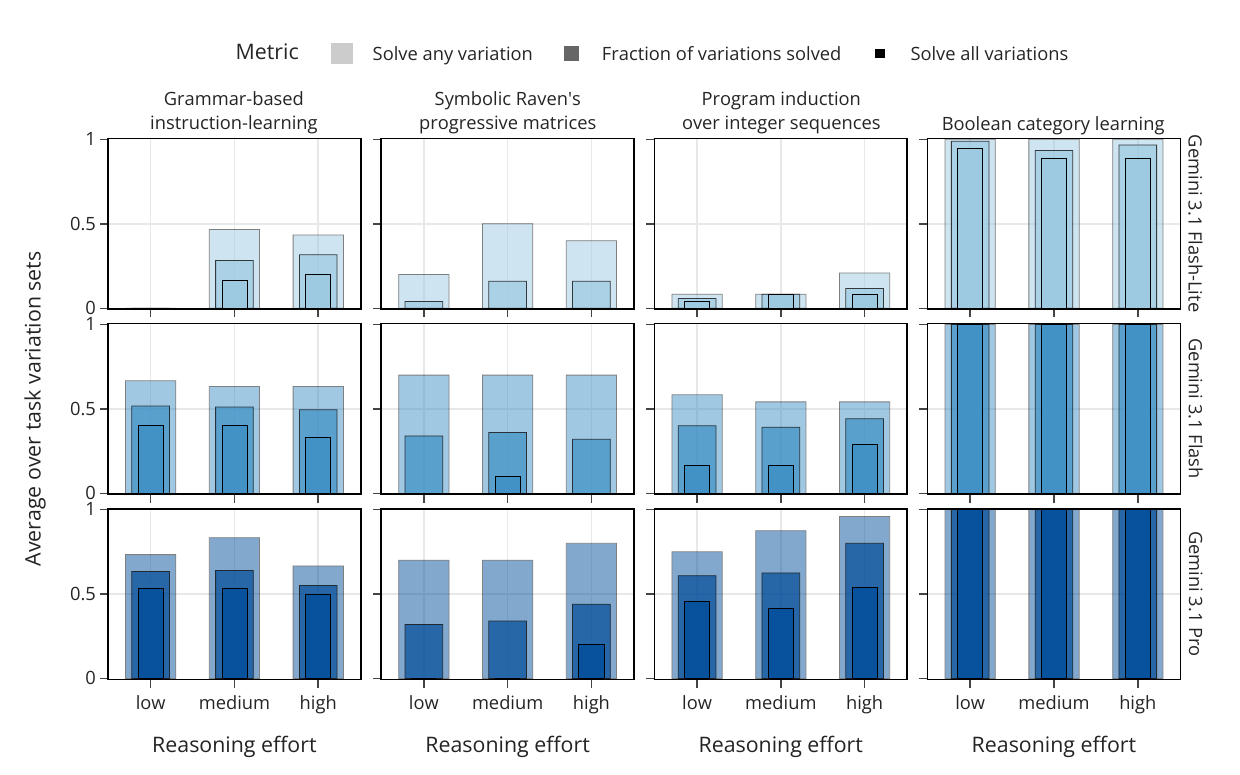}
  \caption{\textbf{Reasoning effort does not consistently improve systematicity.} Varying the reasoning effort for different Gemini reasoning models across task families does not consistently improve systematicity.}
  \label{fig:reasoning-effort}
\end{center}
\end{figure}

\subsection{Reasoning effort does not consistently improve systematicity}
Inference-time scaling with increased reasoning budgets typically improves model performance.
While our main comparison in Figure~\ref{fig:systematicity-gap} was conducted with reasoning effort set to high, we would like to understand the impact of the reasoning effort on systematicity.
Apriori, one might expect that increasing the reasoning effort could reduce the impact of task variations on model variance and therefore possibly improve systematicity.

Figure~\ref{fig:reasoning-effort} shows the systematicity metrics across task families for Gemini models when varying the reasoning effort from \texttt{low} to \texttt{medium}, and \texttt{high}.
Surprisingly, increasing reasoning effort does not consistently improve systematicity.
On the \textit{program induction over integer sequences} and \textit{symbolic Raven's progressive matrices} task families, we can observe a modest improvement in task performance but the systematicity gap largely remains unaffected.
On the \textit{grammar-based instruction-learning} task family the effect is highly model-dependent. 

\subsection{Model stochasticity only partially explains the systematicity gap}
\label{sec:results-stochasticity}
Since reasoning models rely on stochastic sampling, two sources of variance affect model systematicity: the within-task-variation sampling noise and the between-task-variation variance due to the isomorphic task variations.
In the following, we try to delineate the impact of either, first by decomposing the empirically observed variances and then by re-evaluating models with zero temperature sampling to eliminate sampling noise (up to numerical stability).

\paragraph{Large variance for multiple attempts on the same task.}
By applying the law of total variance and using suitable empirical estimators, we can decompose the total variance across task variations and attempts for a given task into the within-task-variation variance and the between-task-variation variance.
In Appendix~\ref{sec:variance-decomposition}, we formally state this decomposition and develop the respective estimators.
Table~\ref{tab:variance_decomposition} lists the different variance components across task families, type of task variation and model.
We find that within-task-variation variance often makes up the largest part of the total variance implying that large sampling noise dominates (or rather masks as we will see in the next section) between-task-variation variance.

\paragraph{Systematicity gap persists with greedy decoding}
\begin{figure}[t]
\begin{center}
    \includegraphics[width=\textwidth]{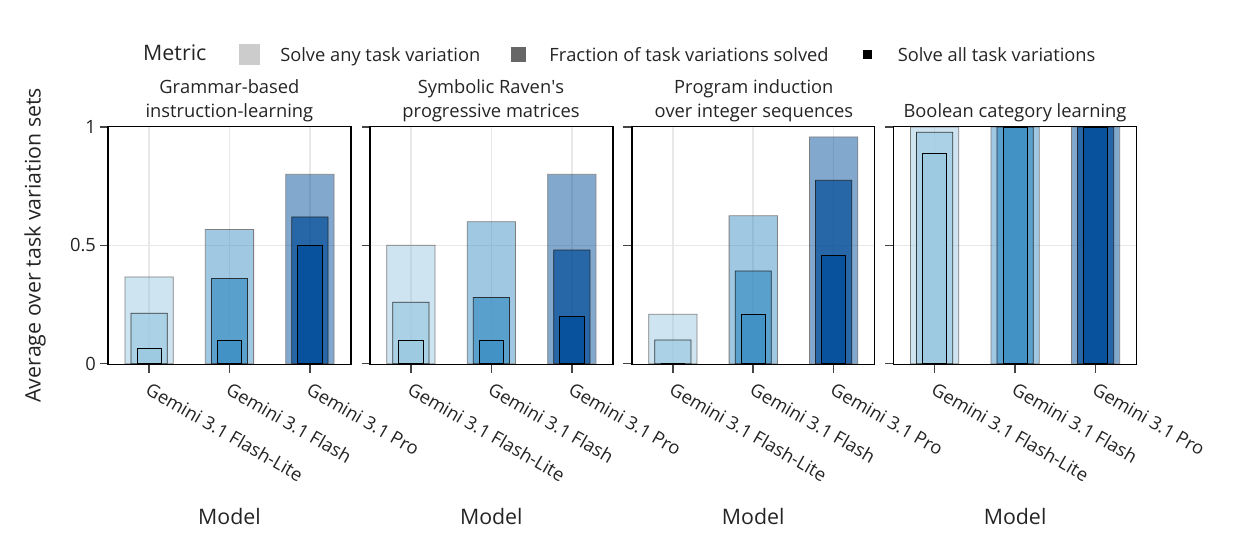}
  \caption{\textbf{Systematicity gap persists with greedy decoding.} When setting the decoding sampling temperature to zero during inference, reasoning model task performance remains comparable and the systematicity gap persists.}
  \label{fig:comparison-greedy}
\end{center}
\end{figure}

By setting the sampling temperature to zero we can, in principle, eliminate the within-task-variation variance.
We re-evaluate Gemini models\footnote{Neither GPT-5.5 nor Claude Opus 4.7 support zero temperature reasoning, so we have to limit this evaluation to Gemini.} in this quasi-deterministic setting\footnote{Floating point arithmetic can lead to residual stochasticity during greedy decoding.}, finding that the systematicity gap and model performance remains largely unchanged, as shown in Figure~\ref{fig:comparison-greedy}.
This suggests that the lack of systematicity observed in reasoning models goes beyond the noisy nature of stochastic decoding.

\section{Discussion}

We investigated systematicity of thought of reasoning models on four rule induction task families -- grammar-based instruction-learning, symbolic reasoning, program induction over integer sequences and Boolean concept learning.
Based on the compositional structure of each task family, we identify task isomorphisms that allow us to generate structurally equivalent task variations.
Across most task families we find that reasoning models lack strict systematicity of thought: despite being able to solve a task, they often fail to solve structurally equivalent variants of it.
These results demonstrate the difficulty of robustly establishing the cognitive abilities of reasoning models beyond the particular contexts they are evaluated in.
Our findings further raise the following points of discussion:

\paragraph{Reliability is a prerequisite for systematicity.}
Consistent behavior when repeatedly facing the exact same context is a requirement for systematicity.
Our findings reveal that even with majority voting over five independent attempts, current reasoning models often do not satisfy this requirement.

\paragraph{Machines should strive to be more systematic than humans.}
We can evaluate systematicity in reasoning models in its strictest form by comparing independent behavioral outcomes across task variations given the exact same initial conditions.
This is not possible with humans where any experience will alter future behavioral outcomes.
Nevertheless, humans are arguably not perfectly systematic either:
Many variables such as mood, tiredness or semantics can affect performance despite having the competence to solve a task in principle \citep{chomsky_aspects_1965, wason_reasoning_1968, evans_conflict_1983, ashby_neuropsychological_1999, van_der_linden_mental_2003}.
In comparison, the intelligent machines we build are tireless and we should strive for them to be as systematic as possible.

\paragraph{Test validity requires systematicity.}
A lack of systematicity raises a fundamental issue of test validity in the evaluation of reasoning models:
Tests have to commit to a particular context and therefore differ by definition from the plethora of contexts within which a system is expected to operate.
Without systematicity, we lack a strong reason to expect that the ability to do something in one context is predictive of this ability in other contexts.

\paragraph{Limitations}
Our evaluation relies on synthetic data generation in order to create the controlled, isomorphic task variations needed to isolate systematicity.
Since measuring systematicity requires evaluating many reasoning chains per task, it is costly and we limit the number of task variations we generate per task to $K=5$ (each of which is evaluated over $N=5$ attempts to compute the majority vote).
Ultimately, a strictly systematic system should solve most task variations, regardless of whether it is evaluated on $K=5$ or $K=100$ variations.

\paragraph{Broader impacts}
This paper evaluates the systematicity of existing reasoning models.
While we foresee no immediate negative societal impact, we hope that it may improve our understanding of this widely deployed technology.

\paragraph{Acknowledgements}
We would like to thank Ryan Burnell, Philipp Mondorf, Changho Shin and Solim LeGris for insightful discussions and valuable feedback.
Simon Schug is supported by Postdoc.Mobility grant \texttt{P500PT\_225369} from the Swiss National Science Foundation.
Brenden M. Lake is supported by the U.S. National Science Foundation (NSF) under Cooperative Agreement No. 2433429, NSF AI Research Institute on Interaction for AI Assistants (ARIA).
We thank Google for generously supporting this research with a financial gift and by providing Google Cloud Credits to evaluate Gemini models.

\newpage
\bibliography{bibliography}
\bibliographystyle{apalike}
\newpage
\appendix

\section{Task variance decomposition}
\label{sec:variance-decomposition}
For each task we sample $K>1$ task variations and evaluate a model's binary success $N>1$ times on each task variation.
This leaves us with two sources of variance: the within-task-variation sampling noise and the between-task-variation variance due to the isomorphic task variations.
Both of these sources can ultimately make the models appear unsystematic.
We quantify their contribution to the total variance in the following.

\subsection{Theoretical quantities}

For a single fixed base task $T$, we define the following random variables:
\begin{itemize}[leftmargin=*]
\item Let $\tau$ be a random variable representing the task variation $\tau(T)$.
\item Let $\hat{p}_\tau$ be the empirical success rate of the model on task variation $\tau$, calculated over $N$ independent attempts.
\item Let $p_\tau := \E[\hat{p}_\tau | \tau]$ be the true underlying success probability of the model on task variation $\tau$.
\end{itemize}

To understand the total variance in the observed success rates, $\hat{p}_\tau$, across all possible task variations, we apply the law of total variance:
\begin{align}
\label{eq:total-variance}
    \Var(\hat{p}_\tau) &= \E_\tau[\Var(\hat{p}_\tau | \tau)] + \Var_\tau(\E[\hat{p}_\tau | \tau])
\end{align}

Because $\hat{p}_\tau$ is the sample mean of $N$ independent Bernoulli trials, each with success probability $p_\tau$, the variance of the empirical success rate given the task variation is $\Var(\hat{p}_\tau | \tau) = \frac{p_\tau(1-p_\tau)}{N}$.
Substituting Equation~\eqref{eq:total-variance} and rearranging the terms yields:
\begin{align*}
    \Var_\tau(p_\tau) &= \Var(\hat{p}_\tau) - \E_\tau\left[ \frac{p_\tau(1-p_\tau)}{N} \right].
\end{align*}
This equation allows us to isolate the true between-task-variation variance of the underlying success probabilities, $\Var_\tau(p_\tau)$, from the expected within-task-variation variance purely due to the finite sampling noise of $N$ attempts per task variation, $\E_\tau\left[ \frac{p_\tau(1-p_\tau)}{N} \right]$ using the total variation of the empirical success rate, $\Var(\hat{p}_\tau)$.

\subsection{Empirical estimates}

For a given task, we sample $K$ task variations.
Let $\hat{p}_{\tau_k}$ be the model's empirical success rate for sampled variation $\tau_k$ (where $k \in \{1, \dots, K\}$), and $\hat{p}$ be the model's overall empirical success rate for this task across all $K \cdot N$ attempts.

We estimate the between-task-variation variance, $\hat{\sigma}^2_\text{between}$, of the underlying success probabilities across variations by removing the estimated within-task-variation variance, $\hat{\sigma}^2_\text{within}$ from the estimated total variance, $\hat{\sigma}^2_\text{total}$, as outlined in the derivation above:
\begin{align*}
    \hat{\sigma}^2_\text{between} &= \hat{\sigma}^2_\text{total} - \hat{\sigma}^2_\text{within}
\end{align*}

We compute the following empirical estimators for the variance components:
\paragraph{Total observed variance} We estimate the sample variance of the variation success rates as:
\begin{align*}
    \hat{\sigma}^2_\text{total} &= \frac{1}{K-1} \sum_{k=1}^K (\hat{p}_{\tau_k} - \hat{p})^2
\end{align*}
\paragraph{Expected within-task-variation variance}
To prevent true variation differences from inflating the estimated noise, we use the following unbiased local estimator to calculate the expected binomial variance over $N$ attempts,
\begin{align*}
    \hat{\sigma}^2_\text{within} &= \frac{1}{K} \sum_{k=1}^K \frac{\hat{p}_{\tau_k}(1-\hat{p}_{\tau_k})}{N-1}.
\end{align*}

\paragraph{Normalization}
Since the empirical success rates differ between tasks, so do the theoretically obtainable maximum variances, which makes a comparison difficult.
For this reason, we normalize the three variance components, expressing them as a proportion of the maximum possible variance.
With each $\hat{p}_{\tau_k} \in [0,1]$, the total observed sample variance is bounded by a theoretical maximum of $\sigma^2_{max} = \frac{K}{K-1}\hat{p}(1-\hat{p})$.
We then obtain the
\begin{itemize}[leftmargin=*]
    \item \text{Normalized total variance}, $\rho_\text{total} = \frac{\hat{\sigma}^2_{\text{total}}}{\sigma^2_{\text{max}}}$,
    \item \text{Normalized within-task-variation variance}, $\rho_\text{within} = \frac{\hat{\sigma}^2_{\text{within}}}{\sigma^2_{\text{max}}}$,
    \item \text{Normalized between-task-variation variance}, $\rho_\text{between} = \frac{ \hat{\sigma}^2_{\text{between}}}{\sigma^2_{\text{max}}}$.
\end{itemize}

\subsection{Results}

In Table~\ref{tab:variance_decomposition}, we report the average normalized variance components across task families, type of task variation and model.
We find that the within-task-variation variance often makes up the largest part of the total variance.

\begin{table}[ht]
\centering
\resizebox{\textwidth}{!}{%
\begin{tabular}{llcccccccc}
\toprule
\textbf{Model} & \textbf{Metric} & \multicolumn{8}{c}{\textbf{Task family}} \\
\cmidrule(lr){3-10}
 &  & \multicolumn{2}{c}{Integer list} & \multicolumn{3}{c}{Grammar-based instruction} & \multicolumn{1}{c}{Raven} & \multicolumn{2}{c}{Boolean concept} \\
\cmidrule(lr){3-4} \cmidrule(lr){5-7} \cmidrule(lr){8-8} \cmidrule(lr){9-10}
 &  & Substitution & Reordering & Recomposition & Reordering & Substitution & Rebinding & Substitution & Reordering \\
\midrule
\multirow{3}{*}{Claude Opus 4.7} & $\rho_{\text{total}}$ & 0.20 & 0.17 & 0.12 & 0.17 & 0.20 & 0.62 & - & - \\
 & $\rho_{\text{within}}$ & 0.16 & 0.17 & 0.18 & 0.17 & 0.16 & 0.08 & - & - \\
 & $\rho_{\text{between}}$ & 0.04 & 0.00 & 0.00 & 0.01 & 0.04 & 0.54 & - & - \\
\midrule
\multirow{3}{*}{GPT-5.5} & $\rho_{\text{total}}$ & 0.19 & 0.15 & 0.17 & 0.17 & - & 0.19 & - & - \\
 & $\rho_{\text{within}}$ & 0.16 & 0.17 & 0.17 & 0.17 & - & 0.16 & - & - \\
 & $\rho_{\text{between}}$ & 0.03 & 0.00 & 0.00 & 0.00 & - & 0.03 & - & - \\
\midrule
\multirow{3}{*}{Gemini 3.1 Flash} & $\rho_{\text{total}}$ & 0.15 & 0.18 & 0.10 & 0.10 & 0.30 & 0.70 & 0.17 & 0.24 \\
 & $\rho_{\text{within}}$ & 0.17 & 0.16 & 0.19 & 0.19 & 0.15 & 0.06 & 0.17 & 0.15 \\
 & $\rho_{\text{between}}$ & 0.00 & 0.01 & 0.00 & 0.00 & 0.15 & 0.64 & 0.00 & 0.09 \\
\midrule
\multirow{3}{*}{Gemini 3.1 Flash-Lite} & $\rho_{\text{total}}$ & 0.22 & 0.36 & 0.11 & 0.23 & 0.05 & 0.37 & 0.19 & 0.15 \\
 & $\rho_{\text{within}}$ & 0.16 & 0.13 & 0.19 & 0.16 & 0.20 & 0.13 & 0.16 & 0.17 \\
 & $\rho_{\text{between}}$ & 0.07 & 0.23 & 0.00 & 0.07 & 0.00 & 0.25 & 0.03 & 0.00 \\
\midrule
\multirow{3}{*}{Gemini 3.1 Pro} & $\rho_{\text{total}}$ & 0.20 & 0.15 & 0.14 & 0.31 & 0.20 & 0.44 & - & - \\
 & $\rho_{\text{within}}$ & 0.16 & 0.17 & 0.18 & 0.14 & 0.17 & 0.11 & - & - \\
 & $\rho_{\text{between}}$ & 0.04 & 0.00 & 0.00 & 0.17 & 0.03 & 0.32 & - & - \\
\bottomrule
\end{tabular}
}
\vspace{1em}
\caption{\textbf{Variance decomposition across tasks and models.} Average estimated variance components for each model and task family grouped by the specific task variation type. Since our estimators are noisy, $\rho_{\text{within}}$ can be larger than $\rho_{\text{total}}$, in which case we clip $\rho_{\text{between}}$ to zero.}
\label{tab:variance_decomposition}
\end{table}

\section{Additional results}

\begin{figure}[t]
\begin{center}
\includegraphics[width=0.5\textwidth]{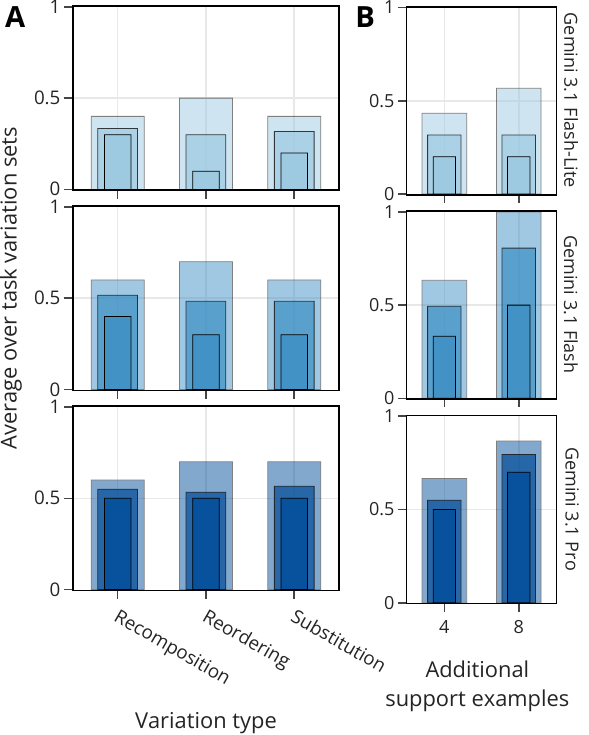}
  \caption{\textbf{Grammar-based instruction-learning.} (A) The systematicity gap persists across variation type. (B) Showing 8 rather than 4 support examples in addition to the curriculum examples increases the average number of tasks where any variation is solved but does not resolve the systematicity gap.}
  \label{fig:mlc}
\end{center}
\end{figure}

\begin{figure}[t]
\begin{center}
  \includegraphics[width=0.8\textwidth]{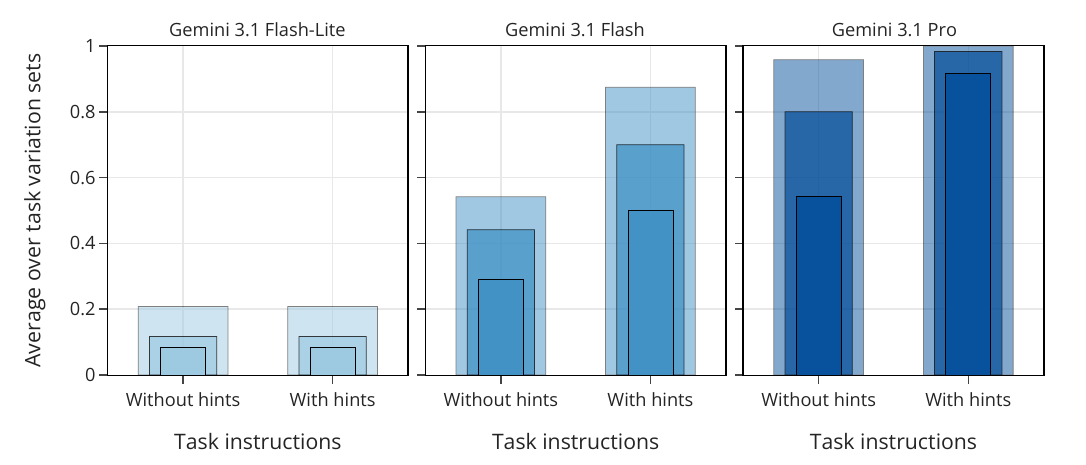}
  \caption{\textbf{Program induction over integer sequences.} Providing hints in the task instructions improves performance for Gemini 3.1 Flash and Gemini 3.1 Pro and almost eliminates the systematicity gap in the latter.}
  \label{fig:listint}
\end{center}
\end{figure}

\begin{figure}[t]
\begin{center}
  \includegraphics[width=0.66\textwidth]{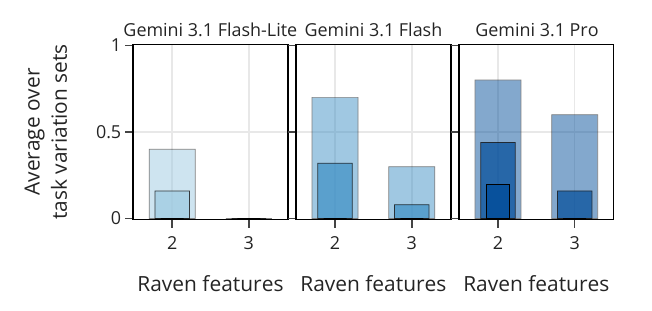}
  \caption{\textbf{Symbolic Raven's progressive matrices.} Increasing the number of features notably increases difficulty and leads to a corresponding drop in the average number of tasks where any variation can be solved.}
  \label{fig:raven}
\end{center}
\end{figure}

Figure~\ref{fig:mlc} shows additional results on the grammar-based instruction learning task family, Figure~\ref{fig:listint} shows additional results on the program induction over integer sequences task family, and Figure~\ref{fig:raven} shows additional results on the Symbolic Raven's progressive matrices task family, all described in Section~\ref{sec:systematicity-evaluation} of the main text.

\section{Task instructions}
For each task family, we provide detailed instructions in the system prompt.
Figure~\ref{fig:mlc-instructions} lists the instructions for the grammar-based instruction-learning task family, Figure~\ref{fig:raven-instructions} lists the instructions for the symbolic Raven’s progressive matrices task family, and Figure~\ref{fig:listint-instructions} lists the instructions for the program induction over integer sequences task family.

\begin{figure}[h]
\begin{taskinstructions}
Your task is to discover the grammatical rules of an artificial language.
In the following, you will be presented with examples of input and output string pairs.
These pairs demonstrate how sequences of primitive tokens are transformed by grammatical rules.
Afterwards, you will be presented with {n\_query} new input sequence(s), and you have to translate them into output sequences based on the grammatical rules you have inferred.
Try to identify the simplest deterministic rules that govern these transformations and apply them to the new input sequences.

Every artificial language consists of:
\begin{enumerate}[leftmargin=*]
  \item Primitive rules: Simple one-to-one mappings from an input token to an output token.
  \item Function rules: These operate on one or two arguments. Functions are represented by specific tokens and apply to their adjacent arguments.
  \begin{itemize}[leftmargin=*]
    \item A 1-argument function appears after its argument (e.g., "arg1 func"). A 2-argument function appears between its arguments (e.g., "arg1 func arg2").
    \item Each argument is either a single primitive token or a sequence of tokens.
    \item A function transforms its evaluated arguments into a new sequence by duplicating and/or rearranging them.
    \item When the same 2-argument function is chained with itself, it evaluates left-to-right (e.g., "arg1 func arg2 func arg3" is evaluated as "[arg1 func arg2] func arg3") otherwise functions that accept sequences follow a strict priority order: If Rule A evaluates before Rule B, and Rule B before Rule C, then Rule A always evaluates before Rule C.
  \end{itemize}
\end{enumerate}
\end{taskinstructions}
\caption{Instructions for the grammar-based instruction-learning task family.}
\label{fig:mlc-instructions}
\end{figure}

\begin{figure}[h]
\begin{taskinstructions}
Your task is to predict integers in structured sequences of integers.
Specifically, you will be presented with tasks that consist of three rows with three columns each.
Each panel in this three by three matrix contains {n\_features} feature(s) encoded as integer(s).
Within a row the {n\_features} feature(s) of the last column can be predicted by applying the correct rules to the first two columns.
Within a task, all rows follow the same underlying rules applied to different inputs.

Each row is delimited by ||, each column is delimited by | and the features within a panel are separated by spaces. The last {n\_features} feature(s) in the last panel are masked with {n\_features} question marks, '?'.

\begin{enumerate}[leftmargin=*]
  \item Try to identify which integers in a given task are governed by the same underlying rules.
  \item Try to identify the {n\_features} rule(s) that govern a task.
  \item Consider the following rules/patterns:
  \begin{itemize}[leftmargin=*]
    \item Constant, e.g. the same integer is repeated
    \item Progression, e.g. integers are systematically incremented or decremented by steps of +1, +2, -1, or -2
    \item Modular addition, e.g. two integers prior in the sequence are summed to produce the next integer
    \item Modular subtraction, e.g. two integers prior in the sequence are subtracted to produce the next integer
    \item Maximum/Minimum, e.g. the maximum/minimum of two integers prior in the sequence makes up the next integer
  \end{itemize}
\end{enumerate}
\end{taskinstructions}
\caption{Instructions for the symbolic Raven's progressive matrices task family.}
\label{fig:raven-instructions}
\end{figure}

\begin{figure}[h]
\begin{taskinstructions}
Your task is to discover the rule that transforms input lists of integers into output lists of integers.
In the following, you will be presented with {n\_support} examples of input and output list pairs.
Afterwards, you will be presented with {n\_query} new input list(s) and you have to apply the inferred rule to predict their corresponding output lists.
\end{taskinstructions}
\textit{With hints}
\begin{taskinstructions}
The rule is a composition of the following basic operations:
\begin{itemize}[leftmargin=*]
  \item insert: Insert an element at a specific position.
  \item remove: Remove an element at a specific position.
  \item repeat: Repeat the list until it reaches a specified length.
  \item shift: Shift all elements by a given offset, wrapping around.
  \item swap: Swap the first and second halves of the list.
  \item tail: Keep only a specified number of elements from the end of the list.
\end{itemize}
\end{taskinstructions}
\caption{Instructions for the program induction over integer sequences task family.}
\label{fig:listint-instructions}
\end{figure}

\begin{figure}[h]
\begin{taskinstructions}
Your task is to discover the meaning of 'wudsy', a word in an alien language.
In the following you will be presented with several examples of objects that are either wudsy or not wudsy.
Afterwards you will be presented with {n\_query} new object(s) and you have to determine whether they are either wudsy or not wudsy.
Try to identify the simplest, deterministic rule for what makes objects wudsy or not wudsy and apply it to decide for the new objects.

Each rule is generated from a context-free grammar. The specific grammar and properties of the objects will be provided for each task. The shortest generated rule that explains the examples is the correct one to base the decision on.

Note in particular that this grammar does not contain a not() operation, keep this in mind when trying to identify the "simplest" rule, i.e. the shortest rule that explains all examples that can be expressed in this grammar.

Respond with the boolean value for each of the {n\_query} new object(s) in their presented order according to the rule you identified.
\end{taskinstructions}

\textit{Example task-specific instruction}
\begin{taskinstructions}
Each object has a shape, color and size.\\
The rules are generated from the following context-free grammar:
\begin{align*}
S &\to \texttt{'true'} \mid \texttt{'false'} \mid \texttt{and}(S, S) \mid \texttt{or}(S, S) \mid \texttt{is\_shape}(O, A) \mid \texttt{is\_color}(O, B) \mid \texttt{is\_size}(O, C) \\
O &\to \texttt{'x'} \\
A &\to \texttt{'circle'} \mid \texttt{'triangle'} \mid \texttt{'square'} \\
B &\to \texttt{'blue'} \mid \texttt{'yellow'} \mid \texttt{'green'} \\
C &\to \texttt{'tiny'} \mid \texttt{'small'} \mid \texttt{'large'}
\end{align*}
\end{taskinstructions}
\caption{Instructions for the Boolean concept learning task family.}
\label{fig:boolean-instructions}
\end{figure}

\section{Additional details}
\subsection{Software and libraries}
\label{appsec:software}
For the results obtained in this paper we build on free and open-source software.
We implemented our experiments in Python using NumPy \citep{harris_array_2020} and JAX \citep[][Apache License 2.0]{bradbury_jax_2018}.
We run API models using Mozilla's any-llm library (Apache License 2.0), utilized WandB \citep[][MIT license]{biewald_experiment_2020} to monitor the progress and results of experiments, and Plotly  \citep[][MIT license]{inc_collaborative_2015} for generating the plots.
We use uv for Python project dependency management \citep[][MIT License]{marsh_uv_2024}.

\end{document}